\documentclass[journal]{IEEEtran}
\usepackage{graphicx}
\usepackage[caption=false,font=footnotesize]{subfig}   
\usepackage{url}

\ifCLASSINFOpdf
\else
\fi
\begin{document}
%
\title{SCTD 3.0: Sonar Common Target Detection in the Wild — A Large-Scale, Multi-Scene Dataset from Real Marine Surveys}
%
%
%

\author{Peng Zhang,}

%
%

\markboth{Journal of \LaTeX\ Class Files,~Vol.~13, No.~9, September~2014}%
{Shell \MakeLowercase{\textit{et al.}}: Bare Demo of IEEEtran.cls for Journals}
%



\maketitle

\begin{abstract}
Synthetic Aperture Sonar (SAS) is a core technology for wide-area detection and recognition of small underwater targets. However, publicly available large-scale, high-quality SAS datasets are extremely scarce, which severely hinders the development of data-driven target recognition methods. Existing SAS benchmark datasets are generally small in scale and limited to single acquisition scenarios, making it difficult to reproduce the complex acoustic scattering phenomena, diverse seabed environments, and multi-pose imaging conditions encountered in real-world detection.To fill this gap, this paper introduces SCTD 3.0—a large-scale real-measured dataset for Sonar Common Target Detection in the Wild in natural waters. The dataset cumulatively contains over 10,000 high-quality real SAS image snippets, collected through field measurements using multi-frequency SAS systems operating at 240 kHz, 450 kHz, and other frequency bands. It covers more than ten typical categories of small underwater targets across various real seabed geomorphological environments, comprehensively encompassing complex detection conditions with multiple observation angles, multiple detection ranges, and multiple operating frequency bands.Notably, this paper establishes a rigorous hierarchical annotation protocol that, for the first time, achieves decoupled and independent labeling of targets' intrinsic physical properties, external scene deployment characteristics, and acoustic scattering imaging phenomena—specifically covering intrinsic properties such as target material, geometric shape, and internal structure; external scene characteristics such as burial state and acoustic shadow integrity; and acoustic scattering features such as specular reflection highlights, edge diffraction, and resonance effects. This enables comprehensive and fine-grained characterization of target imaging features. Finally, this paper constructs a multi-task benchmark evaluation framework adapted to object detection, fine-grained classification, and attribute prediction, and conducts a systematic evaluation of mainstream deep learning models under challenging generalization scenarios such as cross-domain, cross-scene, cross-frequency-band, and cross-view settings. The release of SCTD 3.0 is expected to provide a critical data cornerstone for robust underwater target perception research in open-water environments.
\end{abstract}


%
\IEEEpeerreviewmaketitle

\section{Introduction}
%
%
%
%
\IEEEPARstart{U}{nderwater} small targets—such as sunken objects, cables, underwater facilities, and potential threats—pose a critical demand for wide-area precise detection and Automatic Target Recognition (ATR), which has become a key core technology in the fields of marine resource exploitation, underwater infrastructure maintenance, and maritime security protection \cite{ref1, ref2}. Compared with the limitation of conventional Side-Scan Sonar (SSS), whose azimuth resolution degrades significantly with increasing range, Synthetic Aperture Sonar (SAS) constructs an equivalent large-aperture virtual array by coherently integrating multiple pulse echoes along the track direction, thereby achieving range-independent high azimuth resolution imaging \cite{ref3, ref4}. With its unique imaging advantages of long range, wide swath, and high resolution, SAS has become an indispensable high-end tool for small target perception in complex seabed environments \cite{ref5}. In recent years, deep learning-based data-driven algorithms have achieved breakthroughs in the fields of Computer Vision (CV) and Synthetic Aperture Radar (SAR) \cite{ref6, ref7}, injecting new momentum into the intelligent interpretation of SAS imagery \cite{ref8}. However, constrained by the inaccessibility of the underwater environment and the particularity of acoustic imaging, publicly available high-quality real-measured SAS datasets are extremely scarce, severely hindering the evolution and engineering deployment of intelligent SAS-ATR algorithms \cite{ref9}.

The fundamental bottleneck in constructing a large-scale real-measured SAS dataset lies in the inherently inefficient acquisition process and the extremely high cost of data collection dictated by the SAS imaging mechanism. In natural optical or Forward-Looking Sonar (FLS) acquisition, sensors can continuously capture thousands or even tens of thousands of frames at high frame rates. In contrast, SAS imaging is essentially a challenging physical process that "trades time for spatial precision" \cite{ref10}. In actual sea trials, SAS imposes nearly stringent requirements on the attitude stability, track straightness, and sea state conditions of the acquisition platform. Minute phase perturbations caused by high sea states or weak multi-path reverberations can lead to synthetic aperture failure, resulting in image defocusing or even complete acquisition failure \cite{ref11}. The carrier platform must navigate slowly with millimeter-level attitude control, often yielding only a few dozen effective imaging snippets per day. Behind every successful image lies hours of track calibration and environmental adaptation efforts.

Unlike SAR Earth observation, which can achieve regular and controllable multi-view sampling through fixed orbits and pre-planned flight paths, underwater SAS detection faces an angular barrier imposed by the uncontrollable marine environment. In real wild sea areas, the poses of seabed targets are random and cannot be artificially intervened. Coupled with strong current disturbances, undulating seabed topography, and track deviations of Unmanned Surface Vessels (USVs) / Unmanned Underwater Vehicles (UUVs), the azimuth and grazing angles of targets exhibit disordered, non-uniform, and dispersed characteristics. As a result, existing SAS datasets are generally confined to a single frequency band, a single viewing angle, and specific sediment backgrounds, lacking scene diversity. Models perform excellently on data with the same source distribution, yet suffer a cliff-like performance drop when confronted with "in the wild" generalization tests involving cross-region, cross-frequency-band, or variable-view scenarios \cite{ref12}.

Beyond the difficulties in acquisition, the fine-grained annotation of SAS data also faces extremely high professional thresholds and physical ambiguity. Traditional optical image annotation primarily relies on geometric contours and textural cues, whereas SAS images are high-value multi-view acoustic scattering reconstruction maps \cite{ref13}. Affected by sediment reverberation masking and acoustic shadow diffusion, small targets at long ranges are often submerged in noise, and the multi-view synthesis effect can cause nonlinear distortion of scattering centers, making non-professionals highly prone to interpretation biases \cite{ref14}. More critically, existing SAS datasets only provide simplistic annotations in the form of "bounding box + coarse-grained class label," completely neglecting the physical mechanisms underlying acoustic imaging. In real marine environments, the acoustic representation of the same target can undergo drastic changes due to frequency band differences (e.g., the distinct attenuation and penetration characteristics of 240 kHz versus 450 kHz), the degree of sediment burial, the integrity of acoustic shadows, and internal resonance effects. Conversely, objects with different geometric structures may produce remarkably similar strong scattering highlight regions at certain angles. This "same object with different appearances, different objects with the same appearance" acoustic ambiguity phenomenon makes it difficult for traditional purely data-driven pattern recognition algorithms to learn robust physical representations.

To break through the above bottlenecks, this paper introduces SCTD 3.0 (Sonar Common Target Detection in the Wild)—a large-scale, multi-scene benchmark dataset for general sonar target detection in real natural waters. SCTD 3.0 represents a major leap forward from our previous benchmark construction efforts (SCTD 1.0 and 2.0), with contributions mainly reflected in three aspects: First, it is one of the very few large-scale SAS datasets constructed from real maritime data to date, featuring diversity in target types, operating frequency bands, scene environments, and observation geometric parameters. Second, an attribute-based annotation scheme tailored to the sonar imaging mechanism is proposed, breaking through the evaluation limitations of traditional coarse-grained category recognition and supporting richer algorithm assessment. Third, a multi-task, cross-domain comprehensive evaluation framework is established, enabling researchers to compare the performance and robustness of various data-driven methods under a unified benchmark. We believe that SCTD 3.0 can serve as a high-quality public resource to advance research in underwater acoustic vision, data-driven automatic target recognition, and domain generalization learning in marine scenarios. SCTD 3.0 is available at https://github.com/automlresearch/SCTD-3.0.

\section{Related Work}
Datasets serve as the foundation for developing novel target recognition technologies, particularly in the era of deep learning. In stark contrast to the rapid development of large-scale datasets in the visible-light and SAR remote sensing domains in recent years, the quantity and scale of underwater acoustic target detection datasets have grown extremely slowly—a phenomenon largely attributable to the prohibitive cost of marine data acquisition, the closed-source restrictions arising from military sensitivity, and the inherent difficulties in annotating underwater environments.

\textbf{Early test-range datasets (2000s–2010s)} \cite{ref15}. The earliest underwater target acoustic datasets primarily originated from naval test-range experiments conducted by various nations, with the most representative being controlled mine detection experiments carried out by the U.S. Naval Surface Warfare Center (NSWC) and the NATO Undersea Research Centre (NURC). These datasets typically contained a limited variety of mine targets (such as Manta-type and Rockan-type mines), collected in a single scenario of flat sandy seabeds using side-scan sonar or forward-looking sonar at fixed frequency bands and grazing angles. Due to military confidentiality restrictions, the vast majority of these datasets were not made publicly available, rendering them incapable of supporting broad academic research. Although a few subsequently released public versions expanded the range of target types, their sample sizes remained extremely limited (typically only hundreds to a few thousand target snippets), falling far short of what is required for adequately training deep learning models.

\textbf{Simulated datasets}. To compensate for the scarcity of real-measured data, researchers have constructed various simulated datasets based on acoustic scattering physical models. Typical examples involve using ray tracing or finite element methods to perform acoustic simulations of mines and simple geometric shapes (cylinders, spheres, truncated cones), generating synthetic SAS images that contain target highlights and shadows. Such simulated data allow flexible control over imaging parameters and target geometries, yet persistently suffer from the simulation-to-reality domain gap—simulation models typically assume ideal point scatterers or simplified rough-surface scattering models, making it difficult to accurately reproduce the complex scattering characteristics induced by multi-path propagation, the non-stationarity of seabed reverberation, biological fouling on target surfaces, and sound speed profile perturbations in real marine environments. Consequently, detectors trained solely on simulated data often exhibit poor generalization performance when deployed in real-measured scenarios.

\textbf{The SCTD series and other recent datasets (2018–2025)}. In recent years, with the growing application of deep learning in underwater target detection, researchers have begun constructing and releasing acoustic datasets tailored to specific tasks. The SCTD (Sonar Common Target Dataset) series is one of the very few publicly available datasets in this direction. Its earlier versions, v1.0 \cite{ref16} and v2.0 \cite{ref17}, collected side-scan sonar and synthetic aperture sonar images of various underwater man-made targets in limited scenarios. SCTD 1.0 started from scratch, being the first to address the fundamental challenge of the scarcity of sonar image datasets. By gathering publicly available high-quality data and conducting manual screening and annotation of human-shaped, aircraft-shaped, and shipwreck-type targets, this version was the first to validate that deep learning methods can be effectively applied to sonar image analysis. However, this dataset mixed multiple data sources including side-scan sonar, forward-looking sonar, and synthetic aperture sonar, resulting in inconsistent sensor parameters and an inability to reflect the continuity of real-world scenes. To address this, SCTD 2.0 utilized high-resolution Klein side-scan imaging sonar real-measured data, performing manual fine-grained annotation of rectangular bounding boxes and pixel-level segmentation labels.

\textbf{Core limitations of existing datasets}. Based on the above analysis, existing underwater acoustic target detection datasets generally face the following fundamental challenges: (1) Limited scene diversity—the vast majority of datasets cover only flat sandy or muddy bottom scenarios, lacking diverse real marine environments such as rocky substrates, undulating seabed topography, and sea surface reverberation. The non-stationary reverberation and clutter structures in these complex backgrounds pose severe challenges to detector robustness. (2) Limited imaging condition diversity—data are typically collected using only a single frequency band, a single survey track, and a narrow range of grazing angles, failing to reveal the systematic variations in target scattering characteristics as functions of imaging geometry and frequency. (3) Insufficient target diversity—existing datasets are confined to a limited set of simple geometric shapes, lacking target variants with different materials (metallic, fiberglass, plastic), different internal structures (hollow/solid), and different surface treatment conditions, despite the decisive influence of these factors on actual acoustic scattering. (4) Non-standardized evaluation—different datasets adopt different training/testing splits, evaluation metrics, and experimental settings, hindering fair and objective comparisons among different methods. Therefore, establishing a large-scale, multi-dimensional, standardized SAS small-target detection dataset oriented toward real marine "in the wild" scenarios has become an urgent cornerstone for advancing algorithm development in this field.A comparison of existing sonar image datasets is summarized in Table~\ref{tab:1}.

\begin{table*}[htbp]
\centering
\caption{Statistics of sonar image datasets. Categorization by modality and task. ACRONYMS: C (classification), S (segmentation), D (detection), Visual Question Answering (VQA), Image Generation (IG)}
\label{tab:1}
\begin{tabular}{l l l l l p{2.4cm} l l}
\hline
Dataset       & Modality    & Band (kHz) & Classes & Resolution (m) & Img Size                  & Number of Dataset & Supported Tasks \\
\hline
SCTD 1.0/ 2.0 & SAS+SSS+FLS & N/A        & 5       & N/A            & N/A                       & 357               & D               \\
SASSED        & SAS         & N/A        & 5       & N/A            & N/A                       & 129               & S               \\
SAS           & SAS         & N/A        & 2       & N/A            & N/A                       & 611               & C, D            \\
SCTD 3.0      & SAS         & 240/450    & 10      &                & 640$\times$640, Variable  & 10000+            & C, D, S, VQA, IG \\
\hline
\end{tabular}
\end{table*}

\section{Constructing SCTD 3.0}
The goal of SCTD 3.0 is to maximize the diversity of the dataset in terms of target types, scene environments, and sonar imaging parameters, and to construct, through a hierarchical annotation system, a data benchmark that accommodates multi-task requirements including object detection, segmentation, attribute prediction, and image captioning.

\begin{figure*}[htbp]
    \centering
    \includegraphics[width=\linewidth]{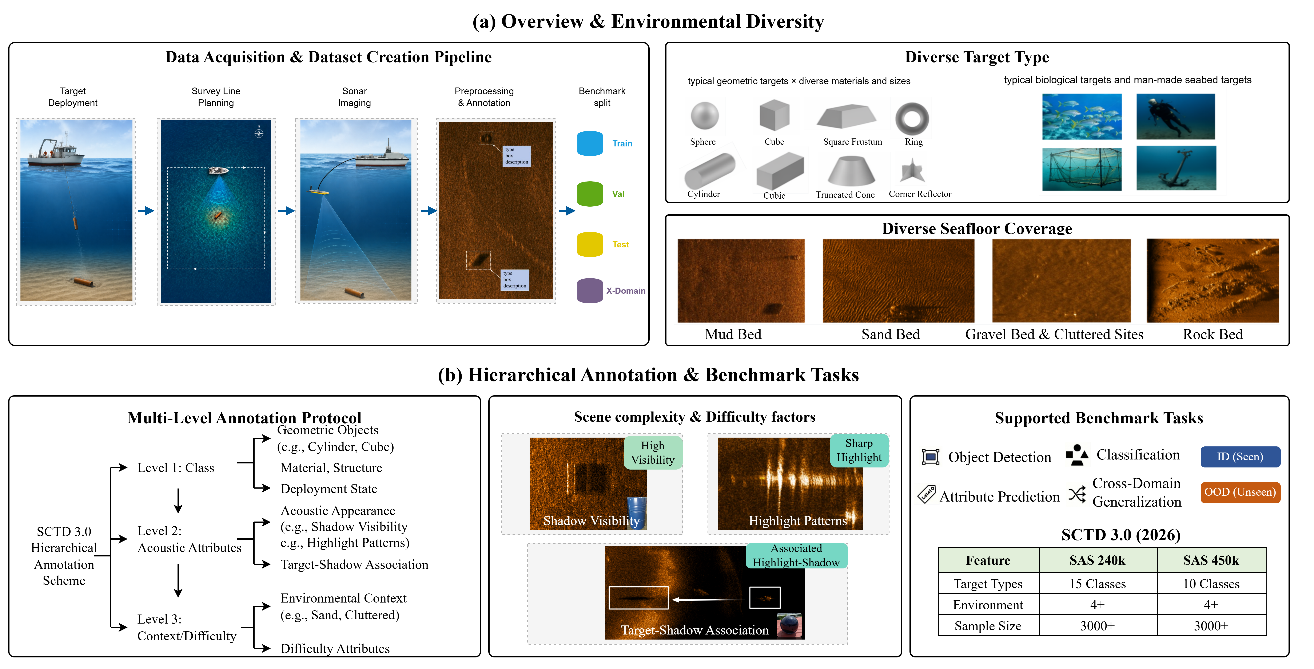}
    \caption{Overview of the proposed SCTD 3.0 dataset.}
    \label{fig:1}
\end{figure*}

As illustrated in Figure~\ref{fig:1}, the construction of the SAS image dataset comprises several stages: data acquisition, image preprocessing, data annotation, and benchmark partitioning. This paper first establishes an SAS image data acquisition pipeline based on unmanned platforms, enabling precise collection of representative SAS images along pre-planned survey tracks in diverse marine environments. Subsequently, multi-faceted annotation is performed on the SAS images of typical targets, innovatively annotating sonar, environmental, and target parameters in a synchronized manner. Finally, to meet the diverse demands of SAS image interpretation, the dataset is rationally partitioned to form benchmarks for multiple categories of tasks.

SAS data acquisition is inherently low in efficiency and constitutes a long-term undertaking. The data collection effort in this work spans multiple sea areas and operational sites, with an overall acquisition period extending over four years (2023 to present).

\subsection{Three Key Elements of Data Acquisition---Targets, Scenes, and Sonar}

The objective of data acquisition is to maximize, through maritime acoustic imaging experiments, the diversity in targets, scenes, and imaging sonar systems, thereby systematically reproducing realistic target detection and recognition scenarios. The target sample set is designed with high diversity, encompassing artificial geometric objects of various geometric forms and material structures, as well as a range of natural and man-made real-world underwater targets. The data acquisition scenes span multiple distinct sea areas, with significant differences in key environmental parameters such as seabed sediment type, detection water depth, and sea surface state across the experimental sites, thereby closely mirroring the complex and variable marine detection environment. The imaging sonar systems and their operational parameters also exhibit diversity, with sonar operating parameters and imaging survey line parameters being differentially configured throughout the experiments.

Target configuration. Underwater small targets exhibit diverse categories and complex morphologies, and fabricating real targets one by one for systematic data collection is extremely costly. To controllably investigate the influence of physical attributes—such as geometric shape, material, and internal filling—on acoustic image characteristics, this study selects regular shapes with typical geometric forms as experimental samples, including ten typical geometric structures: spheres, cylinders, truncated cones, cubes, cuboids, toroids, prisms, corner reflectors, linear objects, and curvilinear objects. Differentiated material and internal filling schemes are assigned to each category of targets. On this basis, to further validate the generalization capability of the dataset in real-world application scenarios, this dataset simultaneously incorporates common irregular targets encountered in actual water-area surveys, including seven typical real-world target categories: small vessels, aquaculture net cages, underwater pile foundations, simulated dummies, divers, fish schools, and sea anchors. This serves to bridge the morphological gap between purely regular-shape samples and real-world underwater targets.

Scenes. To accurately reproduce real-world underwater target detection scenarios using SAS image data, this study conducted data acquisition in actual marine environments, employing multi-configuration imaging sonar systems to collect multiple categories of targets under various imaging conditions and against different marine environmental backgrounds, thereby maximally restoring real marine survey conditions. Seabed sediment type and topographic conditions are key environmental factors that influence the physical properties of underwater targets and their imaging characteristics. Different seabed substrate types can lead to significant differences in the acoustic reflection, imaging texture, and contour intensity characteristics of targets, thereby directly affecting the accuracy and robustness of underwater target recognition and detection. To systematically investigate the imaging characteristics of underwater targets in complex seabed environments and to ensure the representativeness and coverage of the dataset, this study selected four highly representative seabed sediment types in marine environments for real-measured data acquisition, specifically including fine-grained sedimentary mud, medium-grained sedimentary fine sand, coarse-grained sedimentary gravel, and hard-substrate rock, essentially covering the major types of seabed geology.

Sonar. To precisely adjust core acquisition parameters such as imaging perspective and detection range, and to achieve high-fidelity simulation of diverse imaging geometry scenarios, this study employs an unmanned surface vessel (USV) towing a SAS system to conduct imaging acquisition operations in the mission sea areas. This operational approach enables fine-grained control over key parameters, including imaging azimuth angle, the grazing angle of acoustic illumination on targets, and detection range, effectively simulating the complex and variable imaging geometry conditions encountered in real marine survey scenarios, thereby further enhancing the realism and scene adaptability of the collected dataset. To enrich the dimensions of survey conditions and to accommodate the imaging characteristics of different underwater targets, this study utilizes multi-frequency-band SAS for acoustic image acquisition. The differentiated frequency settings can cover operational requirements with varying detection precision and detection ranges. The specific parameters of the SAS systems are presented in Table~\ref{tab:2}.

\begin{table*}[htbp]
\centering
\caption{SAS System Parameter Specifications \cite{ref18}}
\label{tab:2}
\begin{tabular}{lcccc}
\hline
          & Center Frequency (Hz) & Bandwidth (Hz) & Range Resolution (cm) & Azimuth Resolution (cm) \\
\hline
SAS-240k  & 240k                  & 40k            & 2                     & 5                       \\
SAS-450k  & 450k                  & 80k            & 2                     & 3                       \\
\hline
\end{tabular}
\end{table*}

\subsection{Data acquisition}

Selection of survey sites. Several non-sensitive sea areas were identified and selected. The acquisition areas can be marked around wind farm sites, or depicted using maps biased toward underwater/marine resources to illustrate the survey sea-area delineation. Images of the aforementioned USV towing the SAS system are shown in Fig.~\ref{fig:1}.

Beyond the seabed sediment environment, background clutter interference in the oceanic acoustic field is another core variable affecting the imaging quality of underwater targets. To further enrich the scene dimensions of the dataset and enhance the model's adaptability to complex marine interference environments, this study, on top of acquisition across multiple sediment-type scenarios, simultaneously carried out image data collection of underwater targets against sea-surface interference backgrounds and water-column volume reverberation backgrounds. This provides comprehensive and multi-faceted real-measured data support for subsequent research on feature analysis and recognition algorithms for small underwater targets in complex environments.

\subsection{Standardized Data Processing and Annotation}
\subsubsection{Data Organization}

SCTD 3.0 organizes data using a two-level directory structure of "frequency band—scene," where the frequency band primarily corresponds to sonar equipment operating at different frequencies (e.g., 240 kHz and 450 kHz), and the scene corresponds to different sea areas and sediment types. The file directories store acoustic images collected by SAS systems at different frequencies across various survey waters. These acoustic images have undergone subsequent standardized processing and encompass diverse acoustic image characteristics formed by targets in different deployment states and of different types. Consequently, they enable diversified annotation and support a multi-dimensional label system, including target category, target material, orientation angle, imaging clarity, and sediment type.

\subsubsection{Standardization}
To eliminate image interference introduced by the underwater environment and imaging equipment, unify the data input standard, and optimize model detection performance, this study performs standardized preprocessing operations on the raw SAS images, primarily consisting of two core stages: grayscale preprocessing and size cropping preprocessing.

SAS images are uniformly converted to grayscale format, and histogram equalization is applied to enhance target edges and textural details. The necessity of grayscale preprocessing is reflected in two main aspects. On the one hand, during SAS imaging, the echo signal intensity of certain targets may be excessively high, which can easily cause pixel information truncation in the target region, leading to the loss of target features and image distortion. On the other hand, the underwater acoustic propagation mechanism is complex; influenced by physical effects such as water-column attenuation and scattering, the backscattering intensity of the same target exhibits significant differences under different detection ranges and different incidence angles, resulting in non-uniform image grayscale distributions that severely compromise the completeness and consistency of underwater target features.

For size cropping, images are uniformly cropped to 800×800 pixels, with the central region of each image ensured to cover the target body and its typical acoustic shadow, so as to guarantee effective perception of the target's geometric structure and acoustic characteristics during model training. Excessively large image patches lead to redundant computational resource consumption and reduced training efficiency, while excessively small patches can cause the loss of critical acoustic features such as target edges and shadows, or result in structural distortion and shadow truncation, thereby weakening the model's capability for target feature extraction and representation.

\subsubsection{Hierarchical Annotation Framework}

After completing image preprocessing, this study carries out fine-grained, multi-dimensional data annotation. The dataset annotation work primarily implements multi-modal annotation for cooperative target regions and regions of interest that have undergone field verification and validation. The annotation content covers multiple dimensions, including target category, localization information, physical attributes, and imaging feature descriptions. Meanwhile, by differentially partitioning and combining data sources, this study constructs a data benchmark system that can comprehensively support the diverse training, validation, and testing tasks of machine learning models, accommodating the development and performance evaluation requirements of different algorithms for SAS image interpretation.

Target features in SAS images are the result of the coupled interaction among the target's intrinsic physical properties, the seabed environmental background, and underwater acoustic scattering mechanisms. Traditional annotation approaches using a single category label can only achieve coarse target categorization and cannot fully characterize target imaging features in complex underwater acoustic environments, making them ill-suited to meet the training requirements of high-precision, multi-task intelligent underwater target interpretation algorithms. To address this, this paper constructs a hierarchical annotation system based on multi-dimensional attributes, abandoning the single-label definition paradigm and instead comprehensively characterizing target features using fine-grained attribute vectors. Taking a typical underwater target as an example, a cylindrical target with metallic material, hollow structure, partially buried state, and diffuse shadow characteristics can be precisely and completely represented through a multi-attribute tuple of (metal, cylinder, hollow, partially buried, diffuse shadow). This fine-grained annotation paradigm naturally accommodates multi-task learning scenarios and provides data support for training multi-dimensional intelligent interpretation models.

\begin{figure*}[htbp]
    \centering
    \includegraphics[width=\linewidth]{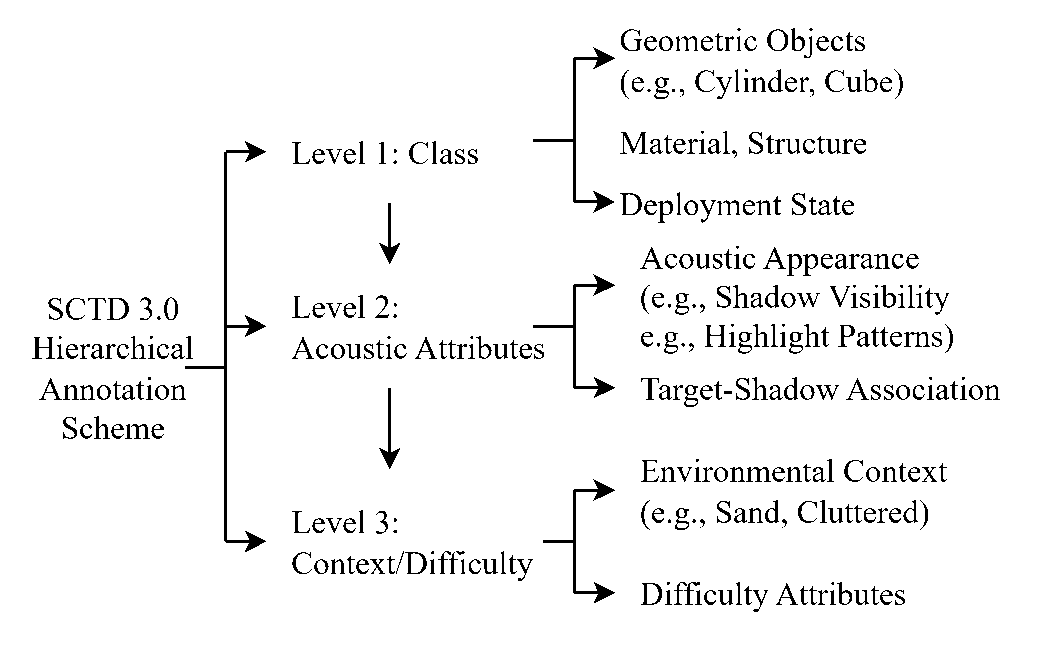}
    \caption{SCTD 3.0 Hierarchical Annotation Scheme.}
    \label{fig:2}
\end{figure*}

Based on the MECE (Mutually Exclusive, Collectively Exhaustive) principle, this study performs fine-grained annotation of all underwater target acoustic image instances across nine independent dimensions, comprehensively covering core elements such as target geometric characteristics, physical attributes, imaging features, deployment states, environmental backgrounds, and recognition difficulties. Specifically, these include:

(1) Geometric shape. Combining the actual morphology of small underwater targets with their acoustic imaging characteristics, target geometric shapes are divided into ten categories, specifically including point-like strong scatterers (e.g., corner reflectors), spheres, cylinders, cones, truncated cones, cubes, cuboids, toroids, linear targets, and irregular targets (e.g., sea anchors, simulated human bodies). Meanwhile, multiple categories of real seabed background samples—such as rocks, clutter, and reverberation—are retained.

(2) Material. Based on the material types commonly used for underwater detection targets, they are categorized into typical materials such as metal, fiberglass-reinforced plastic (FRP), and plastic.

(3) Structure. According to differences in internal structure, targets are classified into solid structures and thin-walled hollow structures.

(4) Highlight region morphology. Based on the imaging characteristics of the target acoustic highlight region, three states are distinguished: well-defined contour, diffuse edges, and no highlight region.

(5) Shadow region morphology. Based on the presentation characteristics of the target acoustic shadow, three types are distinguished: well-defined contour, diffuse edges, and no shadow.

(6) Highlight-shadow association. Characterizes the matching and correlation features between the target's acoustic highlight region and shadow region, depicting the completeness and regularity of target imaging.

(7) Deployment state. Based on the actual distribution state of targets on the seabed, deployment states are categorized as suspended, proud on the seabed, and partially buried.

(8) Background type. Based on the seabed environment of the data acquisition sea areas, backgrounds are classified into muddy bottom, sandy bottom, gravel, and mixed sediment types, covering the mainstream seabed environment types in coastal waters.

(9) Recognition difficulties. Complex marine environments and variable imaging conditions give rise to various image interpretation challenges, which are the core reasons for the failure of underwater target recognition and detection algorithms. Drawing on the complex interference characteristics of "in-the-wild" real-world scenarios in the computer vision domain, the SCTD 3.0 dataset constructed in this paper systematically defines the key difficulty factors in underwater acoustic image interpretation and performs targeted annotation, primarily encompassing six major categories: low signal-to-noise ratio, inconspicuous or absent shadows, reverberation and clutter interference, target burial state, scale variability, and angular variability. The specific characteristics and influence mechanisms of each difficulty factor are as follows: Low signal-to-noise ratio manifests in long-range, high-attenuation detection scenarios, where the saliency of target echo signals is substantially reduced, readily leading to model missed detections. Inconspicuous or absent shadows frequently occur in areas with undulating terrain and in suspended target scenarios, causing the failure of recognition methods that rely on shadow features. Reverberation and clutter interference are primarily induced by hard substrates such as gravel and rocks, as well as water-column volume reverberation; clutter highly overlaps with target features, significantly elevating the algorithm's false alarm rate. The target burial state results in target echo attenuation and incomplete imaging features, leading to target misdetection and missed detection. Scale variability arises from differences in target physical dimensions and variations in slant range, requiring algorithms to possess multi-scale feature extraction capabilities. Angular variability manifests as nonlinear shifts in target scattering characteristics under different detection azimuth and grazing angles, serving as a key factor affecting model generalization capability. The fine-grained annotation of the above difficulty dimensions can not only provide a multi-dimensional testing benchmark for algorithm robustness evaluation but also offer effective feedback for the optimization and iteration of subsequent data acquisition schemes.

Overall, the hierarchical multi-dimensional annotation system constructed in this paper strictly adheres to the MECE principle, breaking through the limitations of traditional single-label annotation. By using multi-dimensional, fully covered attribute vectors to achieve fine-grained characterization of underwater target acoustic images, it can comprehensively support multi-task intelligent interpretation research on small underwater targets in complex scenarios.

For image preprocessing, equalization algorithms, pseudo-color mapping algorithms, etc., are typically applied to perform nonlinear mapping of the original grayscale images. The preprocessing aims to enable annotators to localize targets more clearly and distinguish target features more accurately.

\section{Dataset Statistics}
To date, SCTD 3.0 has collected synthetic aperture sonar images at 240 kHz and 450 kHz. The dataset is partitioned into training, validation, and test sets following a 7:2:1 ratio. Building upon data annotation and benchmark partitioning, SCTD 3.0 supports a range of underwater target recognition tasks, including detection, classification, attribute prediction, and cross-domain detection. Specifically, the detection task focuses on precise localization and bounding-box regression of small targets in complex backgrounds; the classification task requires models to accurately discriminate target categories under low signal-to-noise ratios and strong reverberation interference, as well as to distinguish fine-grained semantics among similar targets (e.g., fish schools versus shipwreck debris); the attribute prediction task demands the accurate output of physical attributes such as target burial depth, orientation angle, and scale ratio; and the cross-domain detection task addresses the generalization requirements across different equipment and water bodies, requiring models to maintain stable detection performance under variations in sonar parameters, water-column sound speed profiles, and sediment-type shifts.
\subsection{SAS-240k}

\begin{figure*}[tb]
\centering
\subfloat[]{\includegraphics[width=0.32\textwidth]{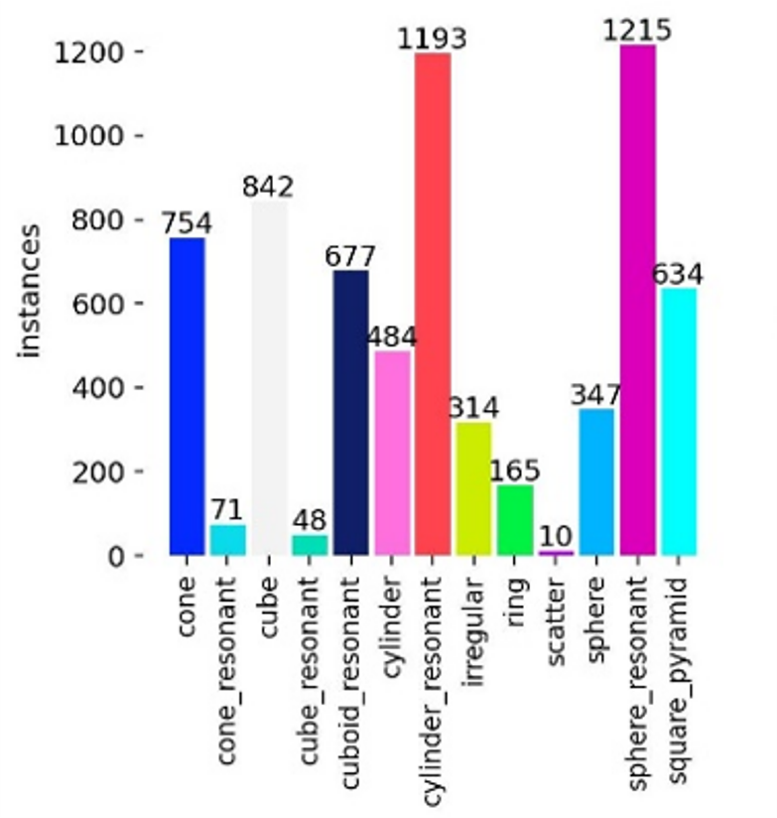}\label{subfig:240a}}
\hfill
\subfloat[]{\includegraphics[width=0.32\textwidth]{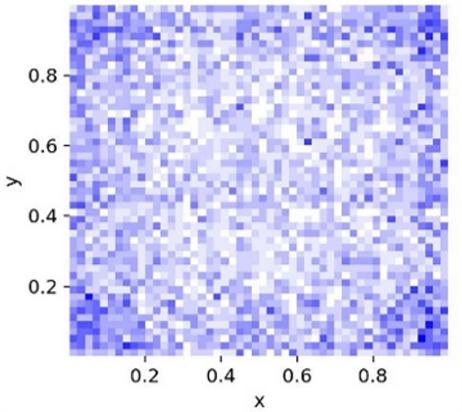}\label{subfig:240b}}
\hfill
\subfloat[]{\includegraphics[width=0.32\textwidth]{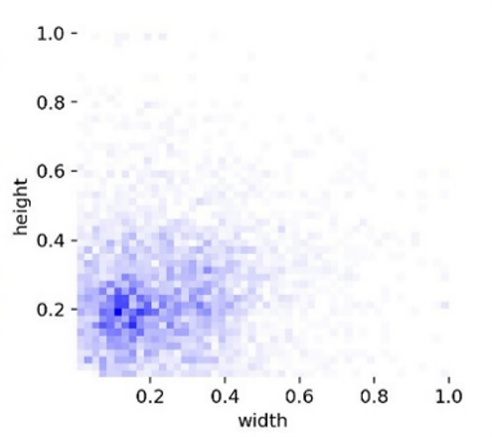}\label{subfig:240c}}
\caption{SCTD 3.0 SAS‑240k sample statistics (as of August 2026).}
\label{fig:3}
\end{figure*}

A total of 13 target categories have been collected and annotated at 240 kHz, namely: ``scatter'', ``cone'', ``cube'', ``cylinder'', ``sphere'', ``ring'', ``irregular'', ``square\_pyramid'', ``cone\_resonant'', ``cube\_resonant'', ``cuboid\_resonant'', ``cylinder\_resonant'', ``sphere\_resonant''. It is worth noting that the dataset includes target variants of the same shape but different materials. Since high-impedance materials such as metals produce stronger scattering centers, the suffix ``\_resonant'' is used to distinguish them. The quantity of each target category and typical examples are shown in the figure~\ref{fig:3}.

The figure above provides a detailed illustration of the sample size distribution across target categories, the spatial distribution of target positions, and the geometric scale characteristics of the bounding boxes in the 240 kHz frequency band. Category imbalance and long-tail distribution.SAS-240k covers 13 typical categories of underwater targets and their resonant states. This non-uniform distribution authentically reproduces the probability differences in the occurrence of specific targets and acoustic phenomena encountered in marine surveys. Spatial and scale characterization.The spatial coordinates (x, y) of targets, both at the image center and periphery, exhibit a uniformly random distribution, thereby eliminating positional prior bias. In terms of geometric scale, the normalized width and height of the vast majority of targets are concentrated around 0.1 × 0.2, indicating that the dataset is dominated by typical small underwater targets.

\subsection{SAS-450k}
The SAS-450k dataset was collected using a 450 kHz high-frequency synthetic aperture sonar system. Compared with the 240 kHz frequency band, the 450 kHz acoustic waves have shorter wavelengths, offering extremely high spatial azimuth resolution and finer sediment texture depiction capability, enabling the clear delineation of geometric edges and subtle acoustic shadow structures of small-sized targets. SAS-450k has collected and annotated a total of 10 typical categories of underwater targets and their physical states, namely: [``scatter'', ``cone'', ``cube'', ``cylinder'', ``sphere'', ``ring'', ``line'', ``irregular'', ``cylinder\_resonant'', ``cone\_resonant''].

\begin{figure*}[tb]
\centering
\subfloat[]{\includegraphics[width=0.32\textwidth]{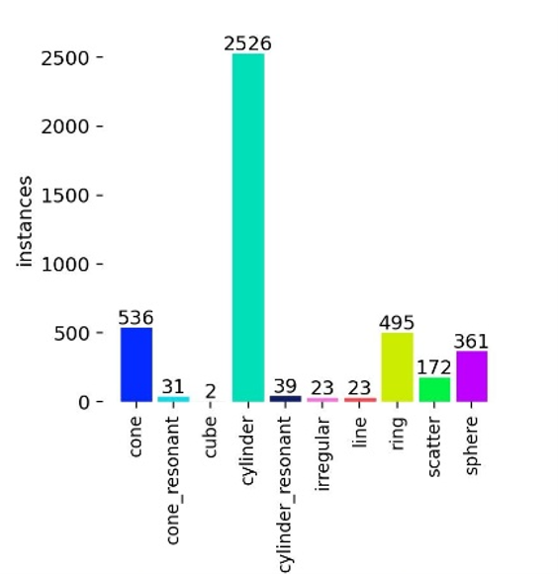}\label{subfig:450a}}
\hfill
\subfloat[]{\includegraphics[width=0.32\textwidth]{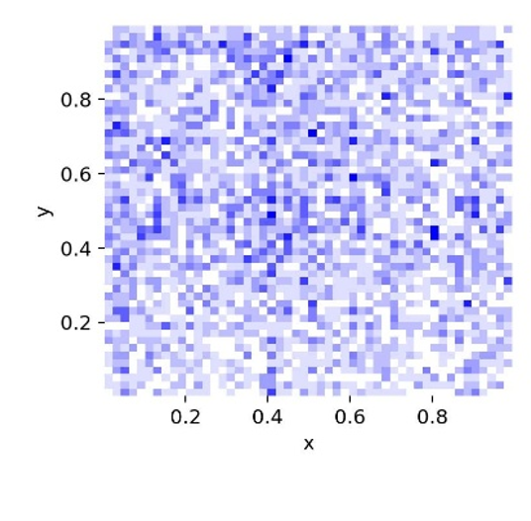}\label{subfig:450b}}
\hfill
\subfloat[]{\includegraphics[width=0.32\textwidth]{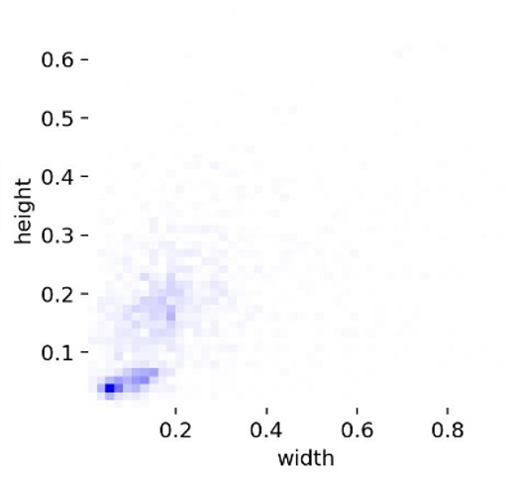}\label{subfig:450c}}
\caption{SCTD 3.0 SAS‑450k sample statistics (as of August 2026).}
\label{fig:4}
\end{figure*}

Consistent with the 240 kHz dataset, SAS-450k also retains the hierarchical decoupled annotation strategy based on acoustic impedance characteristics: for targets where high impedance (e.g., metallic materials) induces strong scattering centers and cavity resonance, the ``\_resonant'' suffix (e.g., cone\_resonant, cylinder\_resonant) is used for independent annotation and differentiation. Meanwhile, leveraging the high-frequency sonar's high discriminability for linear small targets, a linear target category (line, e.g., seabed cables, pipelines) has been newly added. The sample quantity distribution, spatial position distribution, and typical examples of each target category are shown in Fig~\ref{fig:4}.

\section{SCTD 3.0 Applications}
This section evaluates the object detection performance of mainstream algorithms on the SCTD 3.0 dataset.

\subsection{Experiment settings}
Data partitioning. To systematically evaluate the capability of SCTD 3.0 in supporting acoustic image interpretation tasks, this paper constructs a standardized data partitioning scheme covering object detection, instance segmentation, and other tasks, establishing a total of 2 Standard Operating Conditions (SOC) and 3 Extended Operating Conditions (EOC).

Under Standard Operating Conditions (SOC-1, SOC-2), the training and test sets are drawn from the same data distribution.

EOC-1 and EOC-2 respectively validate the transfer scenarios from SAS-240k to SAS-450k and from SAS-450k to SAS-240k. EOC-3 mixes data from both frequency bands in equal proportions before unified partitioning, aiming to investigate whether mixed-frequency-band training can encourage the model to learn frequency-invariant features.

In subsequent work, additional operating conditions will be established to examine the cross-region generalization performance of models across different seabed sediment regions, where the training and test sets are respectively drawn from acquisition areas with different geomorphological types, and to examine model robustness to variations in detection perspective, where the training and test sets are partitioned across different grazing angle intervals. Furthermore, the dataset will be extended to include instance segmentation tasks and attribute prediction tasks.

Considering the differences in available sample sizes in the SCTD 3.0 real-measured data across different frequency bands, regions, and viewing angle conditions, the training/validation/test set ratios and sample selection strategies for each operating condition have been adaptively adjusted according to the actual data distribution. The specific partitioning scheme is shown in Table~\ref{tab:3}.

\begin{table*}[htbp]
\centering
\caption{Summary of Operating Conditions and Data Partitioning.}
\label{tab:3}
\footnotesize
\begin{tabular}{lcccc}
\hline
OC ID & Train Set & Val Set & Test Set & Objective \\
\hline
\multicolumn{5}{c}{\textit{Standard Operating Conditions (SOC)}} \\
\hline
SOC-1 & SAS-240k (70\%) & SAS-240k (15\%) & SAS-240k (15\%) & Baseline under 240k conditions. \\
SOC-2 & SAS-450k (70\%) & SAS-450k (15\%) & SAS-450k (15\%) & Baseline under 450k conditions. \\
\hline
\multicolumn{5}{c}{\textit{Extended Operating Conditions (EOC)}} \\
\hline
EOC-1 & SAS-240k (100\%) & SAS-240k (split) & SAS-450k (100\%) & 240k $\rightarrow$ 450k generalization. \\
EOC-2 & SAS-450k (100\%) & SAS-450k (split) & SAS-240k (100\%) & 450k $\rightarrow$ 240k generalization. \\
EOC-3 & 240k (70\%) + 450k (70\%) & 240k (15\%) + 450k (15\%) & 240k (15\%), 450k (15\%) & Mixed-band domain-invariant feature learning. \\
\hline
\multicolumn{5}{p{0.95\textwidth}}{\footnotesize Note: In EOC-1 and EOC-2, testing is conducted only on common target categories shared between the two frequency bands. Validation sets are split from the corresponding training sets.} \\
\end{tabular}
\end{table*}

Evaluation Metrics. For the object detection task, the COCO standard evaluation metrics are adopted, reporting the mean Average Precision (mAP) over IoU thresholds ranging from 0.5 to 0.95. Additionally, mAP@0.5 and mAP@0.75 are reported to quantify the model's localization accuracy. The F1 Score is also employed to evaluate model performance. This section evaluates mainstream detection algorithms using the Ultralytics YOLO series, with the latest YOLOv26 model selected. All models are trained within the Ultralytics framework, initialized with COCO pre-trained weights. The batch size is set to 16, and the input image resolution is uniformly resized to 800 × 800 pixels.

\subsection{Object Detection}
\subsubsection{Detection on SAS-240k}
\begin{figure*}[htbp]
    \centering
    \includegraphics[width=0.85\textwidth]{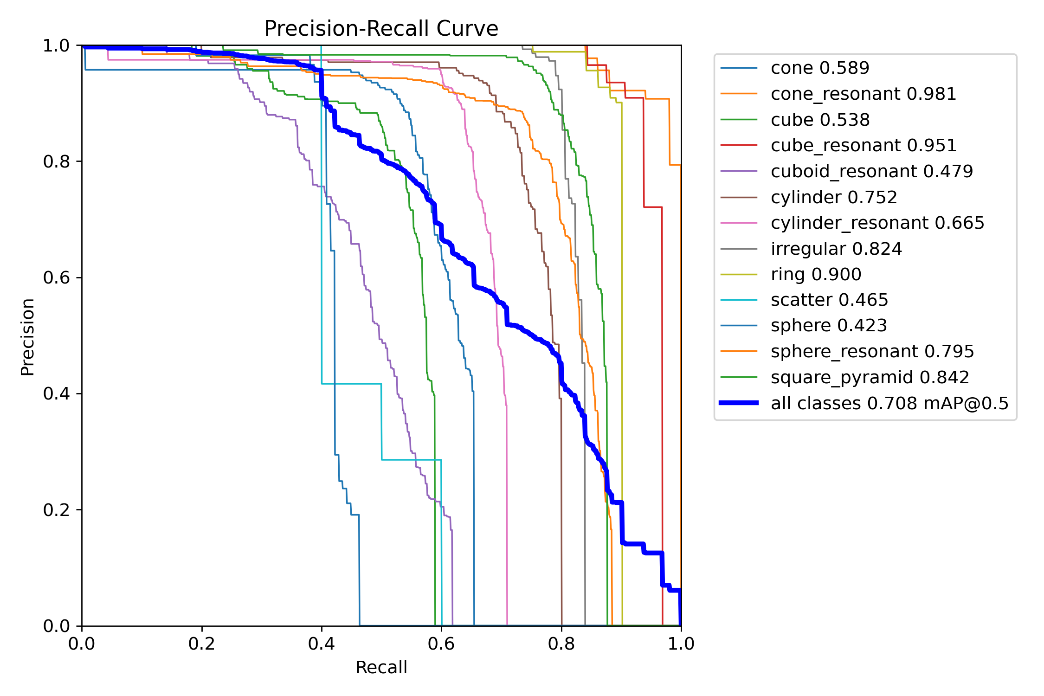}
    \vspace{0.1cm}
    \\ \small (a)
    \vspace{0.3cm}
    
    \includegraphics[width=0.85\textwidth]{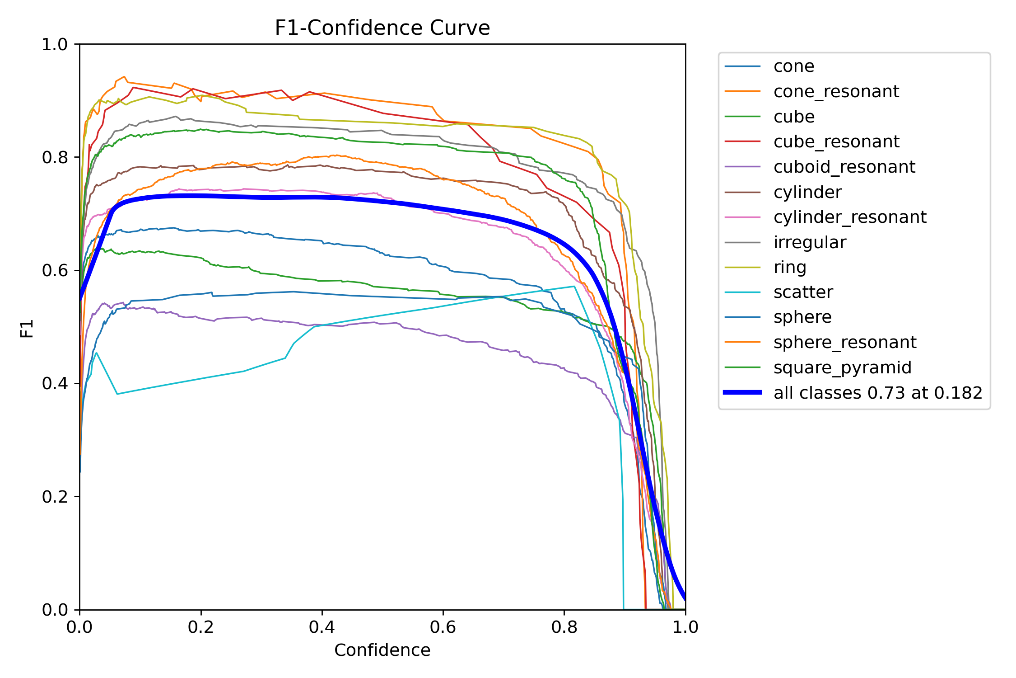}
    \vspace{0.1cm}
    \\ \small (b)
    \caption{Detection results of the baseline model on SAS-240k.}
    \label{fig:5}
\end{figure*}

The quantitative evaluation results of the YOLOv26-based baseline model on the SAS-240k dataset are presented below. The model achieves an overall detection accuracy of 0.708 mAP@0.5. Specifically, Figure ~\ref{fig:5}(a) shows the precision-recall (PR) curve of the model, where the area enclosed by the curve intuitively indicates that the baseline model possesses excellent overall detection performance. However, significant differences exist in the detection difficulty across different target categories within the dataset, with per-class mAP values spanning a wide range from 0.423 (sphere) to 0.981 (cone\_resonant). Figure ~\ref{fig:5}(b) illustrates the comprehensive detection performance score of the model. At a confidence threshold of 0.182, the model achieves an optimal F1 composite response score of 0.73.

Combined with the analysis of the experimental results in light of the hierarchical decoupled annotation characteristics of the SCTD 3.0 dataset, it is further revealed that the physical structure, geometric shape, and acoustic resonance effects of targets are the core factors determining the difficulty of underwater target detection for deep learning models. Among these, acoustic resonance effects exhibit a significant positive contribution to model detection performance, effectively enhancing the detectability and interpretability of targets. A comparison of detection results between ordinary geometric targets and acoustically resonant targets reveals a clear physical pattern: the mAP detection accuracy of resonant-state targets is significantly higher than that of non-resonant-state targets, indicating that acoustic resonance features can serve as effective discriminative cues for underwater target detection.

Differences in target geometric morphology also lead to pronounced divergence in detection performance. The ring target and the square pyramid target exhibit excellent detection performance, achieving mAP values of 0.900 and 0.842, respectively. The underlying reason is that the ring target's distinctive "central hollow shadow" feature and the square pyramid target's unique "multi-faceted corner reflection effect" provide the model with highly discriminative geometric cues, substantially reducing the detection difficulty. In contrast, the sphere target yields the worst detection performance, with an mAP of merely 0.423. Spheres possess isotropic specular scattering characteristics, with diffuse and blurred acoustic shadow edges. Moreover, in complex seabed environments with sand ripple undulations, spheres are prone to sediment burial, making their acoustic highlight features easily confused with random seabed scatterers (scatter, mAP=0.465), ultimately leading to a significant degradation in the model's detection accuracy for sphere targets.

Supplementary experimental results, including the training convergence curves on the SAS-240k dataset, actual detection examples, and the category confusion matrix, are presented in Appendix A (Figs.~\ref{fig:7}, \ref{fig:8}, and \ref{fig:9}).

\subsubsection{Detection on SAS-450k}
\begin{figure*}[htbp]
    \centering
    \includegraphics[width=0.85\textwidth]{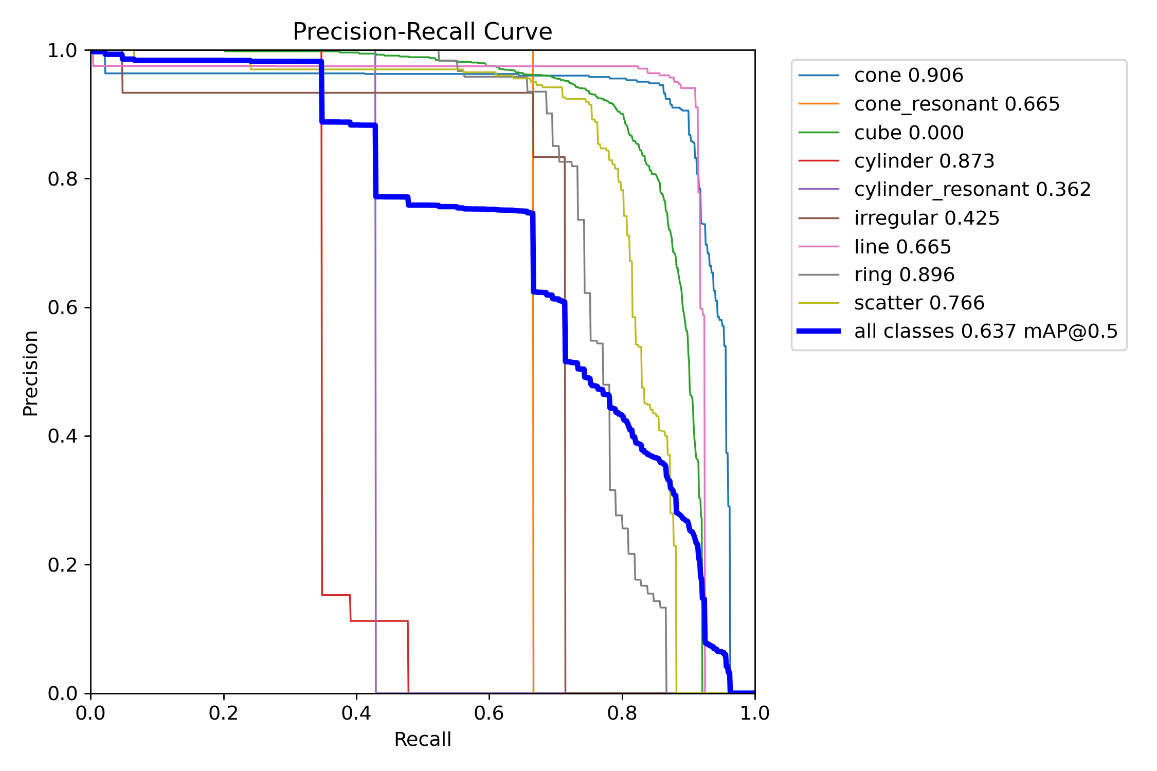}
    \vspace{0.1cm}
    \\ \small (a)
    \vspace{0.3cm}
    
    \includegraphics[width=0.85\textwidth]{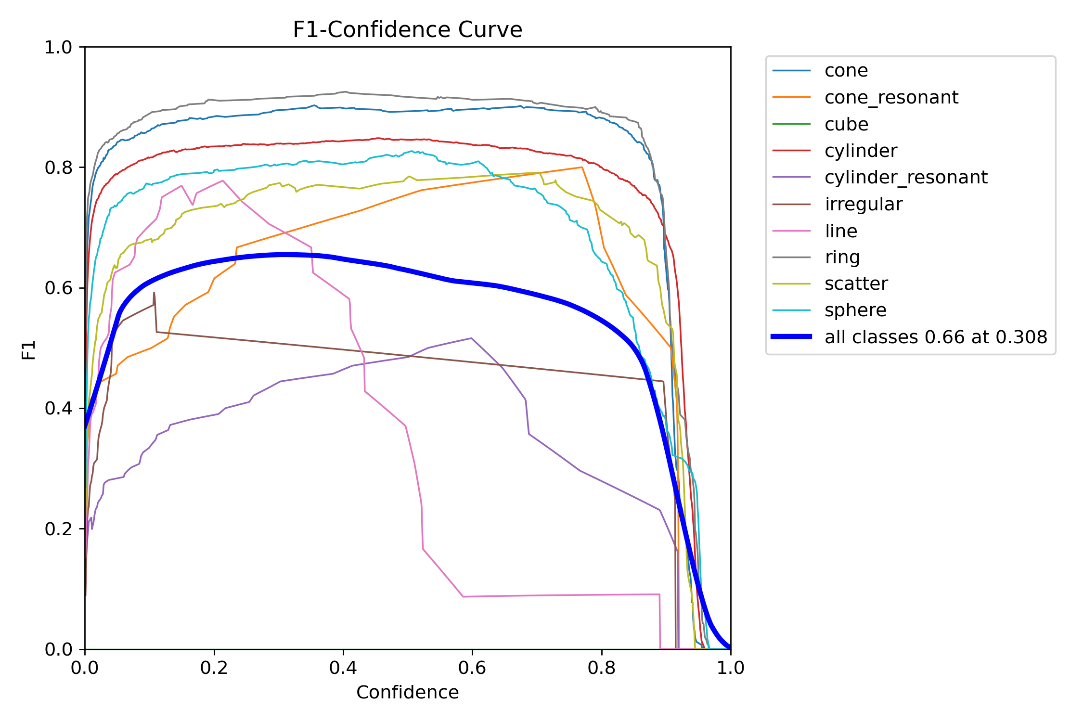}
    \vspace{0.1cm}
    \\ \small (b)
    \caption{Detection results of the baseline model on SAS-450k.}
    \label{fig:6}
\end{figure*}

Compared with the low-frequency band, under 450 kHz high-frequency imaging, the mAP detection accuracy of ordinary geometric targets is overall higher than that of resonant-state targets. For instance, the cone target achieves an mAP of 0.906, significantly outperforming the cone\_resonant target (mAP=0.665); the cylinder target achieves an mAP of 0.873, also substantially surpassing the cylinder\_resonant target (mAP=0.362). Investigating the underlying physical and data mechanisms, the 450 kHz high-frequency SAS possesses extremely high spatial resolution, capable of precisely delineating the clear specular highlight and geometric shadow contours of conventional geometric shapes. However, resonant acoustic signals are susceptible to medium absorption attenuation and sediment reverberation interference during high-frequency propagation. Coupled with the extreme scarcity of resonant-state samples (e.g., only 31 instances of cone\_resonant and 39 instances of cylinder\_resonant), the model's generalization capability for resonant textures is constrained. The confusion matrix, as shown in Fig.~\ref{fig:6}, reveals that 52\% of the true cylinder\_resonant samples are erroneously predicted as background.

Target geometric morphology and sample size similarly induce pronounced divergence in detection performance. The cone, ring, and cylinder targets exhibit the most outstanding detection results, achieving mAP values of 0.906, 0.896, and 0.873, respectively. This is attributable to the distinctive "central hollow shadow" feature of the ring target and the strong "bright spot—long acoustic shadow" combination of the cone and cylinder targets in high-resolution 450 kHz sonar images, which provide the model with highly discriminative geometric visual cues, substantially reducing the detection difficulty.

In contrast, the cube target, due to an extreme lack of training samples (only 2 instances), fails to enable the model to effectively learn its features, yielding an mAP of merely 0.000 (the confusion matrix reveals 50\% misclassification as cylinder and 50\% misclassification as line). The irregular target (irregular, mAP=0.425), lacking fixed shape characteristics and exhibiting diffuse and blurred shadow edges, results in up to 57\% of true samples being overwhelmed by complex seabed sediments (predicted as background). Furthermore, the discrete scatter target (scatter), with 172 instances, achieves an mAP of 0.766, indicating that high-frequency SAS possesses a favorable capability for capturing small strong scatterers on the seabed.

Supplementary experimental results on the SAS-450k dataset, including training convergence curves, recall/precision response curves, sample geometric distributions, and the category-normalized confusion matrix, are presented in Appendix B (Figs.~\ref{fig:10}, \ref{fig:11}, and \ref{fig:12}).

\section{Discussion and Future Work}
To address the critical bottleneck that current publicly available benchmark datasets for underwater Synthetic Aperture Sonar (SAS) are limited in scale, confined to single scenarios, and lacking in physical interpretability—thereby severely constraining the generalization capability of data-driven underwater Automatic Target Recognition (ATR) algorithms—this paper introduces SCTD 3.0, a large-scale, multi-scene benchmark dataset for general sonar target detection tailored to real-world marine engineering environments.

The main contributions and core findings of this paper are summarized as follows. (a) Breakthrough in data scale and real-world operating conditions: SCTD 3.0 aggregates over 10,000 high-resolution real SAS image snippets sourced from actual marine surveys, spanning mainstream operating frequency bands including 240 kHz and 450 kHz. The dataset not only covers more than 10 typical categories of small underwater targets, but also comprehensively reproduces real acoustic imaging characteristics under multiple detection perspectives, multiple detection ranges, and complex seabed geomorphology, thereby filling the gap in large-scale, in-the-wild SAS standard benchmarks. (b) Physically decoupled hierarchical annotation system: This paper is the first to propose and apply a tripartite hierarchical decoupled annotation protocol encompassing "intrinsic physical properties—external scene characteristics—acoustic scattering phenomena." This not only enables precise characterization of target geometry, burial state, and shadow features, but also provides robust annotation support for physically interpretable deep learning research grounded in acoustic mechanisms. (c) Multi-task benchmarking and revelation of generalization bottlenecks: A composite benchmark evaluation framework covering object detection, fine-grained classification, and attribute prediction is constructed. Systematic evaluations reveal that mainstream data-driven models still suffer significant performance degradation when confronted with challenging domain shift tasks such as cross-frequency-band, cross-scene, and cross-view scenarios, highlighting the unique value of SCTD 3.0 as a high-difficulty generalization testbed.

As a critical data cornerstone for underwater acoustic perception in open waters, SCTD 3.0 will strongly propel the research community toward overcoming domain generalization and few-shot learning challenges in real marine environments. In future work, we will deepen and extend this research along the following dimensions: first, the continuous expansion of the data ecosystem; and second, the exploration of physically driven ATR paradigms.


%

\section*{Acknowledgment}

\ifCLASSOPTIONcaptionsoff
  \newpage
\fi



%

\appendix

\section{SAS-240k Detection Results}

\begin{figure*}[htbp]
    \centering
    \includegraphics[width=\linewidth]{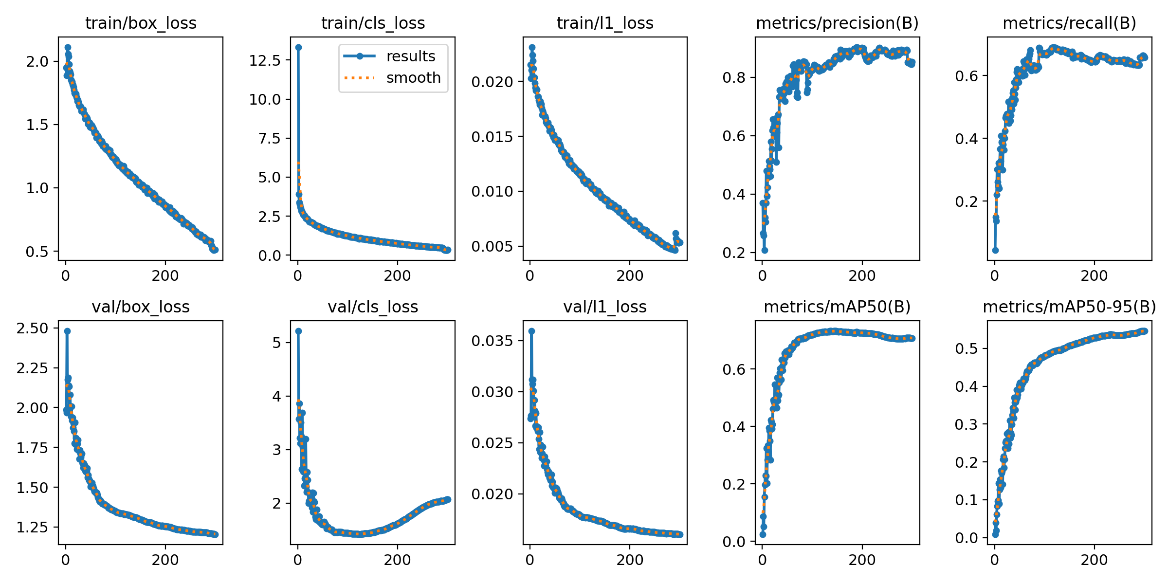}
    \caption{Training curves on the SAS-240k dataset.}
    \label{fig:7}
\end{figure*}
\begin{figure*}[htbp]
    \centering
    \includegraphics[width=0.5\linewidth]{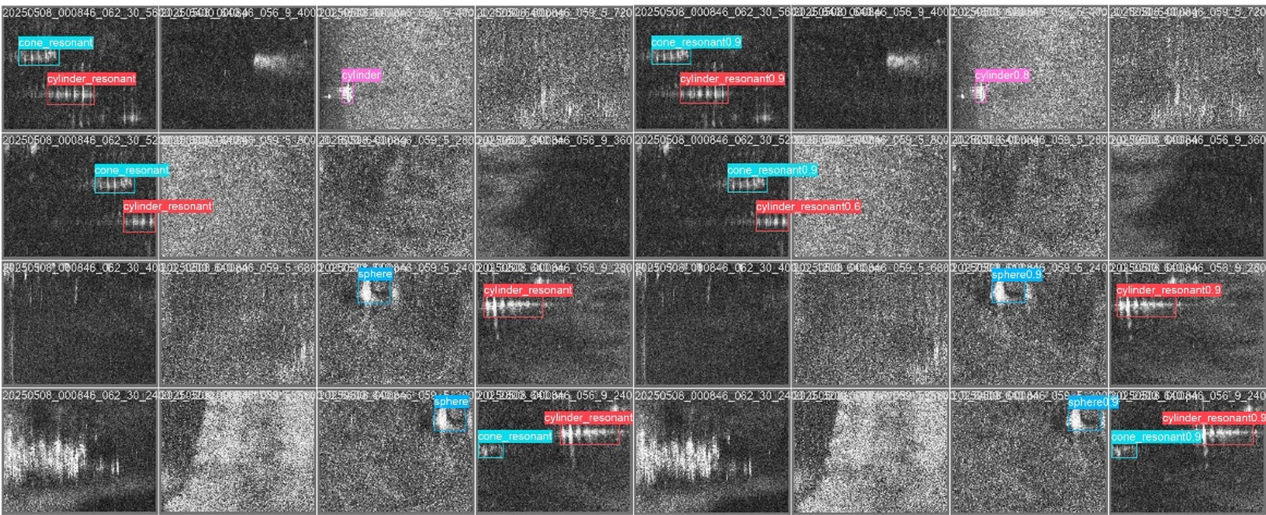}
    \caption{Typical detection results on SAS-240k.}
    \label{fig:8}
\end{figure*}
\begin{figure*}[htbp]
    \centering
    \includegraphics[width=0.5\linewidth]{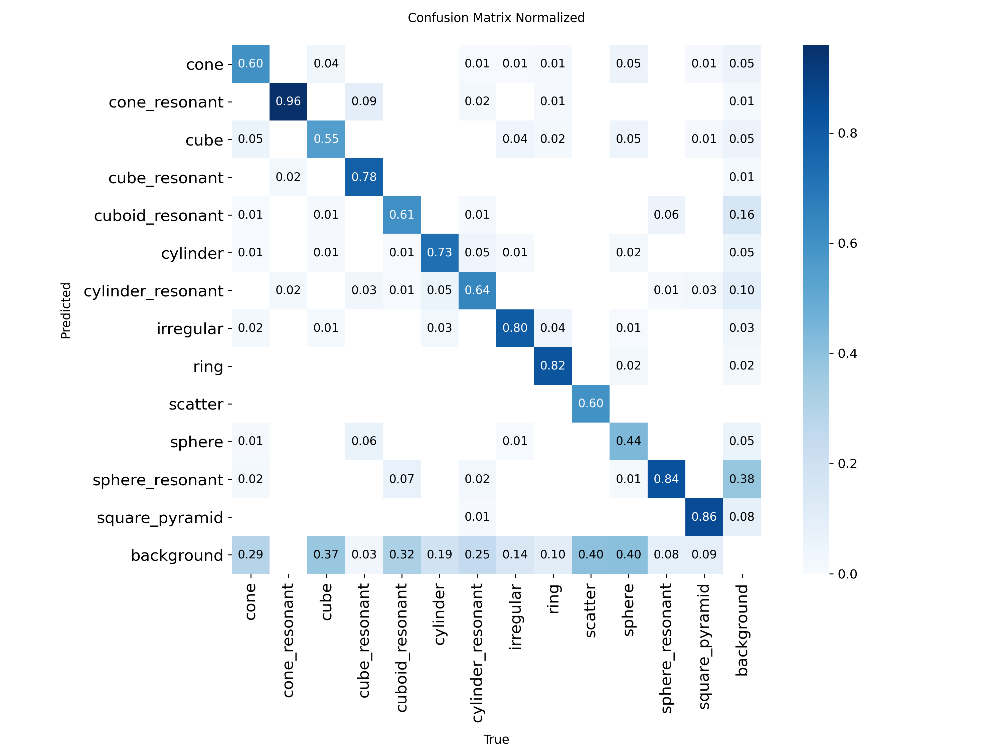}
    \caption{Confusion matrix for target classification on SAS-240k.}
    \label{fig:9}
\end{figure*}

\clearpage

\section{SAS-450k Detection Results}

\begin{figure*}[htbp]
    \centering
    \includegraphics[width=\linewidth]{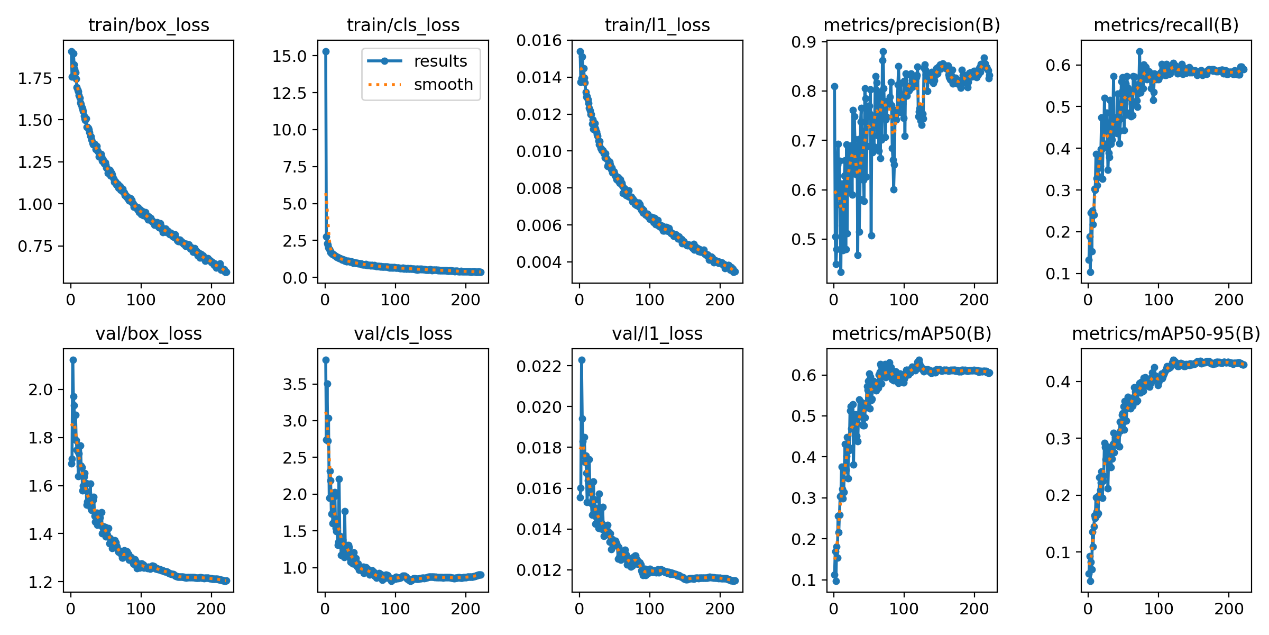}
    \caption{Training curves on the SAS-450k dataset.}
    \label{fig:10}
\end{figure*}
\begin{figure*}[htbp]
    \centering
    \includegraphics[width=0.5\linewidth]{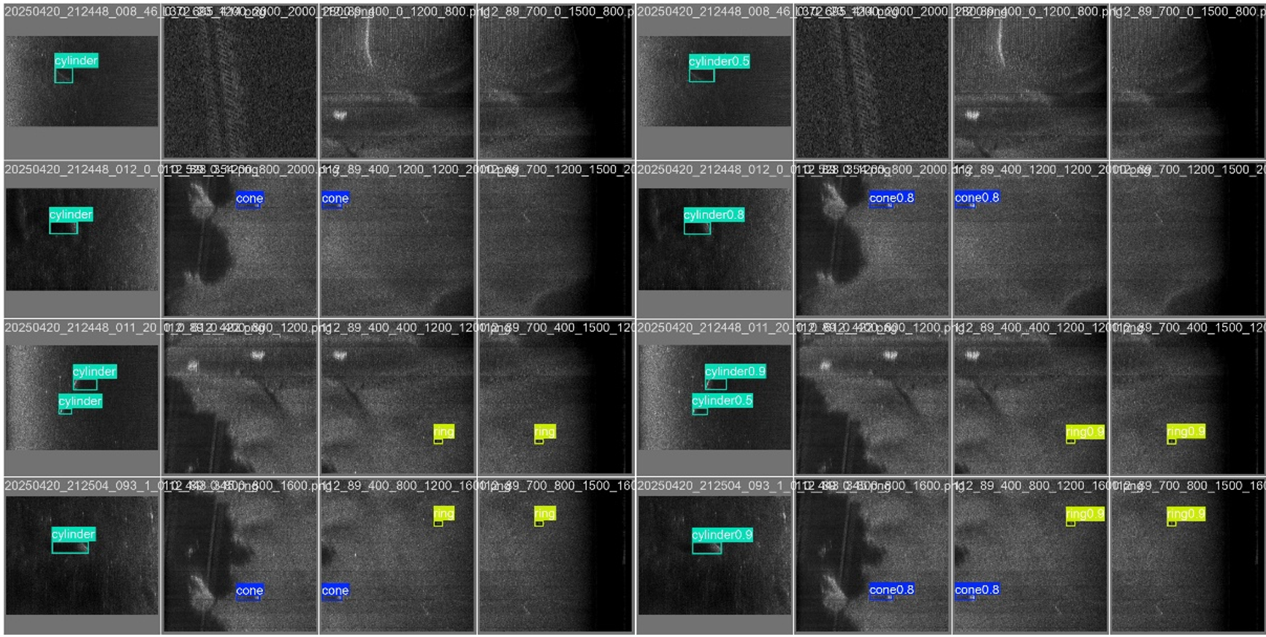}
    \caption{Typical detection results on SAS-450k.}
    \label{fig:11}
\end{figure*}
\begin{figure*}[htbp]
    \centering
    \includegraphics[width=0.5\linewidth]{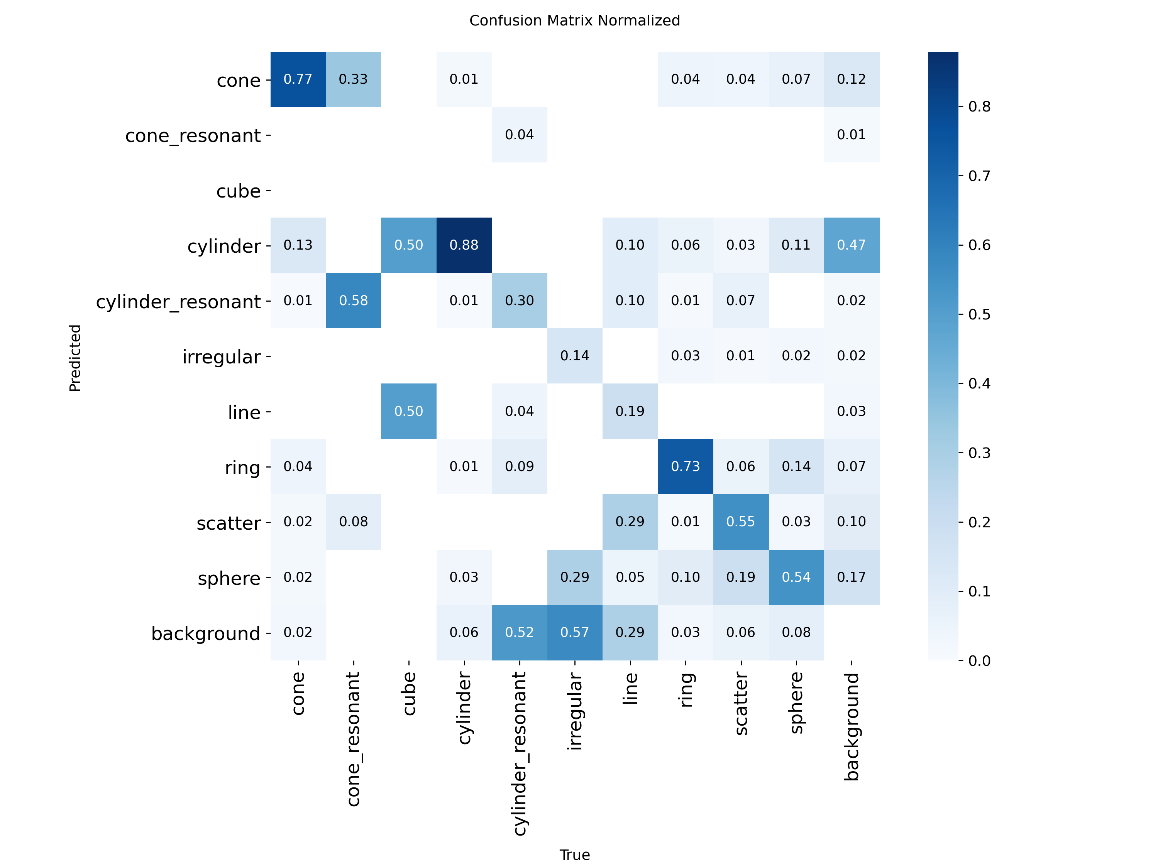}
    \caption{Confusion matrix for target classification on SAS-450k.}
    \label{fig:12}
\end{figure*}

%








\end{document}